%% file: main.tex
\documentclass[10pt,twocolumn,letterpaper]{article}

\usepackage{cvpr}              
\usepackage{multirow}
\usepackage{url}
\usepackage{booktabs}
\usepackage{graphicx}
\usepackage{xcolor}
\usepackage{wrapfig}
\usepackage{subcaption}
\usepackage{caption}

\usepackage{graphicx}
\usepackage{amsmath}
\usepackage{amssymb}
\usepackage{booktabs}
\usepackage{amssymb}
\usepackage{multirow}
\usepackage{multicol}
\usepackage{bm}
\usepackage{colortbl}
\usepackage{wrapfig}
\usepackage{pifont}
 

\definecolor{cvprblue}{rgb}{0.21,0.49,0.74}
\usepackage[pagebackref,breaklinks,colorlinks,allcolors=cvprblue]{hyperref}

\def\paperID{393} 
\def\confName{CVPR}
\def\confYear{2025}

\title{ProxyTransformation: Preshaping Point Cloud Manifold \\ With Proxy Attention For 3D Visual Grounding }

\author{Qihang Peng\\
\\
\and
Henry Zheng\\
 \\
Tsinghua University\\
\and
Gao Huang$\thanks{Corresponding Author.}$
}

\begin{document}
\maketitle
\input{sec/0_abstract}    

\input{sec/1_intro}

\input{sec/2_relatedworks}
\input{sec/4_method}
\input{sec/5_experiment}
\input{sec/6_conclusion}

\section*{Acknowledgement}
The work is supported in part by the National Natural Science Foundation of China under Grant U24B2017, and Beijing Natural Science Foundation under Grant QY24257.

{
    \small
    \bibliographystyle{ieeenat_fullname}
    \bibliography{main}
}

\input{sec/X_suppl}

\end{document}

%% file: sec/0_abstract.tex
\begin{abstract}

Embodied intelligence requires agents to interact with 3D environments in real time based on language instructions. A foundational task in this domain is ego-centric 3D visual grounding. However, the point clouds rendered from RGB-D images retain a large amount of redundant background data and inherent noise, both of which can interfere with the manifold structure of the target regions. Existing point cloud enhancement methods often require a tedious process to improve the manifold, which is not suitable for real-time tasks. We propose \textbf{Proxy Transformation} suitable for multimodal task to efficiently improve the point cloud manifold. Our method first leverages \textbf{Deformable Point Clustering} to identify the point cloud sub-manifolds in target regions. Then, we propose a \textbf{Proxy Attention} module that utilizes multimodal proxies to guide point cloud transformation. Built upon Proxy Attention, we design a submanifold transformation generation module where textual information globally guides translation vectors for different submanifolds, optimizing relative spatial relationships of target regions. Simultaneously, image information guides linear transformations within each submanifold, refining the local point cloud manifold of target regions. Extensive experiments demonstrate that Proxy Transformation significantly outperforms all existing methods, achieving an impressive improvement of \textbf{7.49\%} on easy targets and \textbf{4.60\%} on hard targets, while reducing the computational overhead of attention blocks by \textbf{40.6\%}. These results establish a new SOTA in ego-centric 3D visual grounding, showcasing the effectiveness and robustness of our approach.
\end{abstract}


%% file: sec/1_intro.tex
\section{Introduction}
\label{sec:intro}

\begin{figure*}[t]
    \centering
    \includegraphics[width = \linewidth]{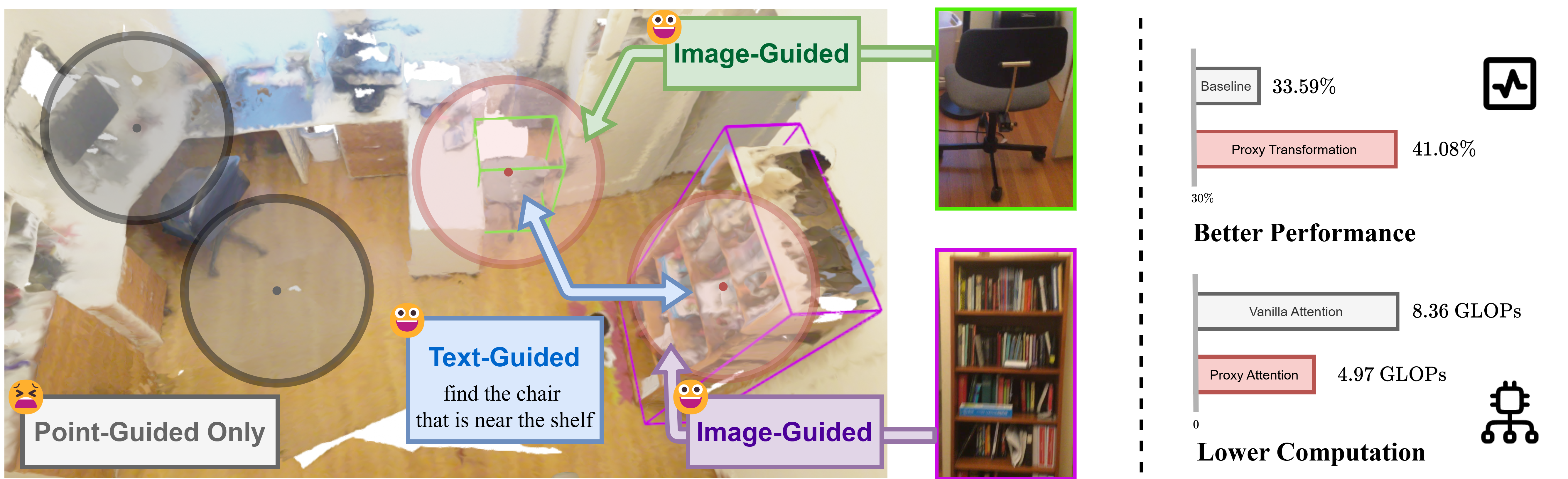}
    \caption{Illustration of our main idea and SOTA results. Ground truth and reference boxes are shown in \textcolor{green}{green} and \textcolor{purple}{purple} respectively. Circular regions represent 3D areas where point cloud enhancement is applied. \textcolor{black}{Black} areas indicate regions that do not contribute to grounding performance and would increase unnecessary computation overhead. Traditional single-modality point cloud guidance would enhance these redundant areas. \textcolor{red}{Red} regions highlight areas where multimodal-guided point cloud enhancement is efficiently applied. Text modality, containing global relative position information among different critical objects, guides translation vectors for these region, while image modality, with local fine-grained semantic details, guides transformation matrices within each target regions. Our model achieves better results with reduced computation, about which details are in~\cref{table:val,tab:attn}.}
    \label{fig:illu}
\end{figure*}

In recent years, the field of embodied AI has gained increasing attention, spurred by 3D visual grounding benchmarks~\citep{achlioptas2020referit3d, chen2020scanrefer, wang2023embodiedscan} that have led to a surge of research~\citep{guo2023viewrefer, wu2023eda, zhao20213dvg, jain2022bottom, yang2024exploiting, huang2022multi, wang2023embodiedscan}. The 3D visual grounding task, which involves locating target objects in real-world 3D environments based on natural language descriptions, is a core perception capability for embodied agents. This ability is crucial in enabling agents to interpret and interact with their surroundings via language, which supports applications in robotics and human-computer interaction.

Despite these advancements, several significant challenges continue to hinder the performance of 3D visual grounding systems. One key challenge is the limited perception of embodied agents, who rely on ego-centric observations from multiple viewpoints, often lacking a holistic, scene-level understanding. While some methods attempt to improve scene-level perception using preconstructed 3D point clouds~\citep{wu2023eda, zhao20213dvg, jain2022bottom, yang2024exploiting, guo2023viewrefer, huang2022multi}, following previous 3D perception approaches~\citep{wu2024pointv3, jiang2020pointgroup}, they are impractical in real-world applications where comprehensive scene-level information is not readily available.

In response, EmbodiedScan~\citep{wang2023embodiedscan} was introduced, utilizing multi-view ego-centric RGB-D scans to enable models to process scenes directly from sparse, partial viewpoints. Previous methods\cite{wang2023embodiedscan, zhu2024scanreasonempowering3dvisual, zhengdenseg,zhengdensegrounding} decouple the encoding of RGB images and depth-reconstructed point clouds from ego-centric perspectives to extract both semantic and geometric information, which are then projected into 3D space using intrinsic and extrinsic matrices to form a semantically enriched point cloud for bounding box regression. While effective, these methods overlook the noise, such as errors in depth on non-lambertian surfaces, captured by the depth sensors \cite{ikemura2024robust, tykkala2011direct, tariqul2017robust} that causes suboptimal performance. Consequently, critical geometric details of corresponding target objects are lost, potentially distorting their original manifold and further compromising the model’s grounding performance. Moreover, a substantial portion of the sampled points represents background regions, leading to a diminished density of foreground object points.

In light of such critical challenges, previous works have focused on improving point cloud structure such as point cloud denoising \cite{wang2023transformer,de2023iterativepfn,de2024straightpcf} or completion methods~\cite{boulch2022poco,leng2024point,zhang2024walkformer, yu2021pointr}. However, these methods often require extensive preprocessing the point cloud data, making them unsuitable for real-time ego-centric 3D visual grounding. Moreover, these methods rely on traditional statistical~\cite{vizzo2022make} methods or learning-based methods~\cite{luo2021score,rakotosaona2020pointcleannet} to enhance the point cloud structure in single modality, which cannot fully utilize the multimodal information available in our task.


Given the limitations of the above methods in ego-centric 3D visual grounding, we propose \textbf{Proxy Transformation}, which enhances point cloud manifolds before feature learning.
This approach effectively reduces redundant computation in background point cloud regions and fully leverages the available multimodal information in current context to optimize the submanifolds of target regions. Notably, our method does not require pre-trained offline reconstruction of scene-level point clouds, making it better suited for real-time ego-centric 3D visual grounding task.

Specifically, to generate transformations for point cloud submanifolds, reference points are first initialized as a uniform 3D grid for different scenes. Motivated by the success of deformable offsets in 2D domain~\cite{xia2022vision,dai2017deformable}, a 3D offset network then takes the initial point cloud clusters centered on these reference points as input, producing offsets for each reference cluster center. In this way, these deformable cluster centers shift toward critical regions. Subsequently, they serve as cluster centers in a succeeding stage to select candidate submanifolds on the original point cloud for further transformations. For each submanifold, we employ a novel generalized attention mechanism, \textbf{Proxy Attention}, to learn the corresponding transformation matrixes and translation vectors. Specific proxy tokens can be selected based on task requirements (\eg downsampling points after pooling, multi-view image features or textual features).


In the context of online point cloud submanifold enhancement, we utilize text and image features as proxy tokens to guide submanifold transformations. In this approach, text information provides global positional relationships among different submanifolds, while image information offers local semantic details within each submanifold, as illustrated in~\cref{fig:illu}. Leveraging these designs, our method supplies the subsequent 3D backbone with a higher-quality point cloud manifold, thereby enhancing the model's effectiveness and robustness.

In summary, our contributions are as follows. (1) We propose Proxy Transformation, enabling multimodal point cloud augmentation in the context of 3D visual grounding. (2) To obtain more desirable submanifolds for target regions, we design deformable point clustering, utilizing a 3D offset network to generate flexible and adaptive deformable clusters suited to diverse scenes. (3) We introduce a generalized proxy attention mechanism, allowing the selection of different proxy tokens based on task requirements, achieving linear computational complexity. (4) Our model significantly outperforms all existing methods, achieving an impressive improvement of \textbf{7.49\%} on easy targets and \textbf{4.60\%} on hard targets, while reducing the computational overhead of attention blocks by \textbf{40.6\%}, establishing a new SOTA in ego-centric 3D visual grounding.

%% file: sec/2_relatedworks.tex
\section{Related Works}
\label{sec:related}

\noindent\textbf{3D Visual Grounding.} 3D visual grounding integrates multimodal data to localize target objects in 3D point clouds and it has gained significant attention in recent years~\citep{dionisio20133d,wang2019reinforced, feng2021cityflow}. 3D visual grounding methods are divided into one-stage and two-stage architectures. One-stage approaches~\citep{liao2020real,luo20223d,geng2024viewinfer3d,he2024refmask3d} fuse textual and visual features to generate predictions in a single step, enabling end-to-end optimization and faster inference. However, they may struggle with complex scene layouts due to the absence of explicit object proposal refinement.In contrast, two-stage approaches~\citep{yang2019dynamic,achlioptas2020referit3d,huang2022multi,guo2023viewrefer,wu2024dora,chang2024mikasa} follow a sequential process: they first utilize pre-trained object detectors~\citep{jiang2020pointgroup,wu2024dora} to generate object proposals, which are subsequently matched with the linguistic input to identify the most likely target. This separation of detection and grounding allows for more precise alignment of visual and textual features, enhancing accuracy, particularly in complex scenes, but at the cost of increased computational overhead and inference time. 

Recent studies~\citep{jain2022bottom,roh2022languagerefer,huang2022multi,Shi_2024_CVPR,chang2024mikasa} have explored transformer-based structures for 3D visual grounding. BUTD-DETR~\citep{jain2022bottom} incorporates outputs from pre-trained detection networks, including predicted bounding boxes and their corresponding class labels, as additional inputs. Multi-View Transformer~\citep{huang2022multi} projects the 3D scene into a multiview space to learn robust feature representations. MiKASA Transformer~\citep{chang2024mikasa} integrates a scene-aware object encoder and an multi-key-anchor technique, enhancing object recognition accuracy and understanding of spatial relationships. In our paper, we focus on the more challenging one-stage methods with transformer modules.  

\noindent\textbf{Point Cloud Enhancement.}
3D point clouds provide rich geometric information and dense semantic details, playing a critical role in various domains~\cite{yang2022ransacs,xiang2023multi,wang2023embodiedscan,yu2021pointr}. However, their inherent sparsity and irregularity limit model performance, and the deficiencies of existing sensors exacerbate issues such as high noise levels, sparsity, and incomplete structures in point clouds~\cite{muzahid2020curvenet,zhu2019vision}. These challenges make 3D point cloud augmentation a critical and challenging problem~\cite{quan2024deep}. Traditional point cloud enhancement methods are often based on interpolation and optimization techniques~\cite{vizzo2022make}, but their computational and memory overhead limits their applicability to large-scale datasets. Current deep learning-based point cloud augmentation methods can be broadly categorized into three main approaches: point cloud denoising, completion, and upsampling.

Point cloud denoising~\cite{zhao2022noise,chen2022repcd,wang2023transformer,de2023iterativepfn,de2024straightpcf} can eliminate noise in the point cloud while approximating and preserving the underlying surface geometry information and semantic details. Point cloud completion focuses mainly on completion for objects~\cite{chen2024learning, leng2024point, yu2021pointr,zhang2024walkformer} and scenes~\cite{xia2023scpnet,xu2023casfusionnet,wang2024semantic,zhang2023point}. And point cloud upsampling~\cite{yu2018pu,liu2022pufa,kumbar2023tp,rong2024repkpu,du2022point} aims to improve the resolution of point clouds while maintaining the integrity of the original geometric structure.

However, these methods are still confined to a single point cloud modality and are not well-suited for tasks with high real-time requirements. In the context of ego-centric 3D visual grounding, multimodal information is available. We leverage both textual and visual modalities as guidance to transform point cloud sub-manifolds. This approach enables simultaneous point cloud denoising and densification, providing higher-quality data for subsequent processing. Additionally, our method is designed with efficiency in mind, ensuring it meets the task's real-time requirements.

%% file: sec/4_method.tex
\section{Methodology}
\label{sec:method}

\begin{figure*}
    \centering
    \includegraphics[width=\linewidth]{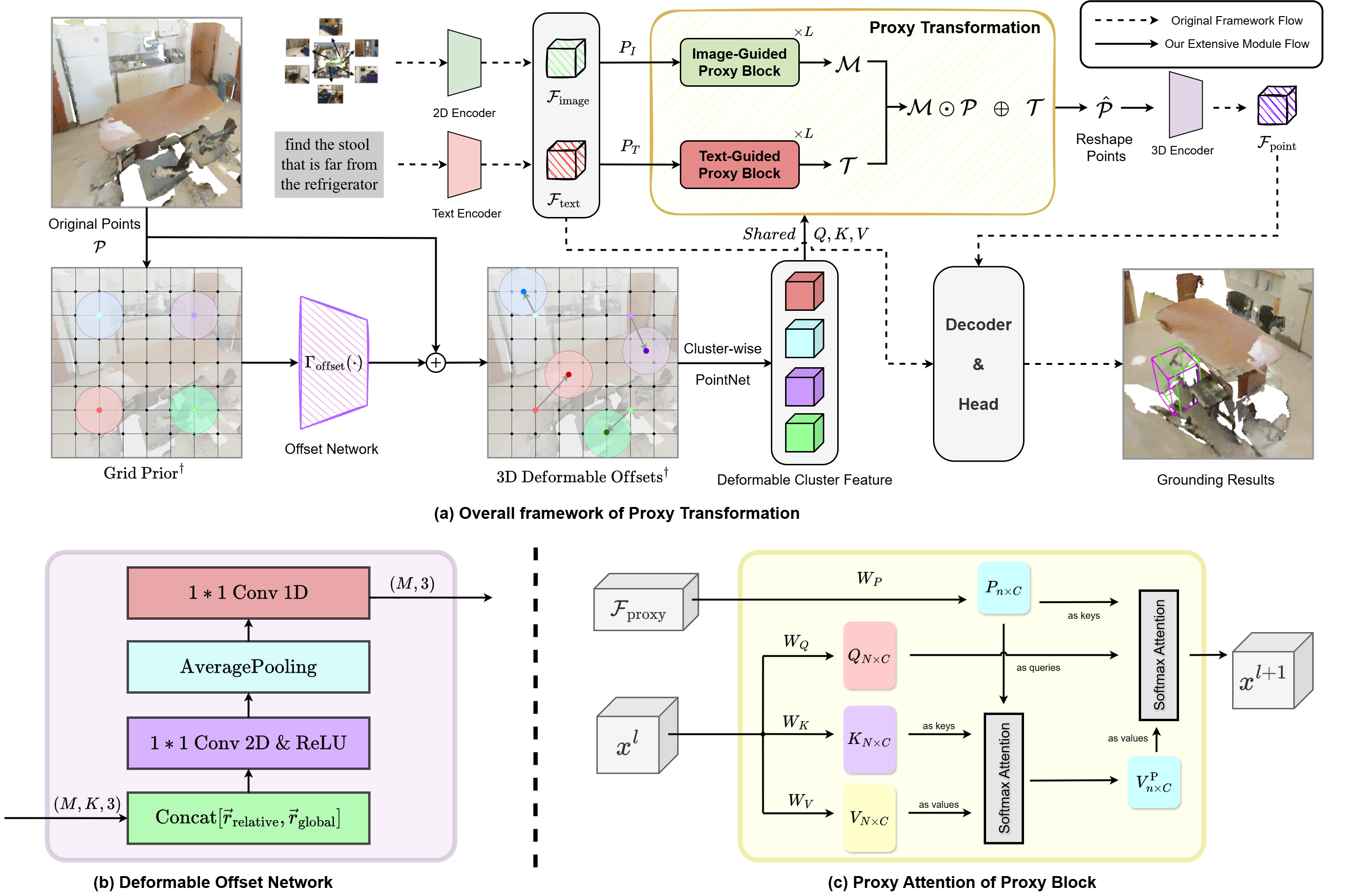}
    \caption{(a) shows the overall framework of Proxy Transforamtion. For simplicity, $^\dagger$ indicates that a 2D grid is used to represent the 3D spatial grid, with generated 3D offsets also expressed as 2D vectors for clarity. The Grid Prior spans the entire space, and we illustrate it with only four reference points for clearer visualization. In Proxy Transformation module, $\mathcal{M}$ and $\mathcal{T}$ are sets of transformation matrixs and translation vectors for all clusters. (b) details the structure of our deformable offset network, and indicates the input and output shapes, where \( M \) represents the number of clusters and \( K \) represents the number of points per cluster. (c) illustrates the information flow in proxy attention. This module combines with FFN and skip connection in a standard Transformer architecture to form the Proxy Block. }
    \label{fig:framework}
\end{figure*}

In this section, we will introduce our network design. Firstly, we will give a brief introduction of the entire pipeline and what our module does in \cref{sub:overall}. Then, we introduce our effective yet efficient Proxy Attention in \cref{sub:proxy} and how deformable offsets are generated for reference points in \cref{sub:deform}. Finally, details of Proxy Transformation will be described in \cref{sub:trans}.

\subsection{Overall Structure}
\label{sub:overall}
We improve upon the previous SOTA in ego-centric 3D visual grounding~\cite{wang2023embodiedscan}. Here we simply review the overall structure. First, the input RGB-D images from each view is transformed into a partial point cloud, which is then integrated into a holistic 3D point cloud \(P \in \mathbb{R}^{N \times 3}\) using global alignment matrices, thereby preserving geometric details. Next, multi-scale image and point features \(\{\mathcal{F}^s_{\rm img}\}^S_{s=1} \in \mathbb{R}^{V \times H_s \times W_s \times C_s}\), \(\{\mathcal{F}^s_{\rm point}\}^S_{s=1} \in \mathbb{R}^{N \times C_s}\), and text features \(\mathcal{T} \in \mathbb{R}^{L \times C}\) are learned by respective backbone. Here, \(N\) is the total number of points sampled from all perspective views \(V\), \(S\) is the number of 
scale for the encoded feature maps and \(L\) is the length of text sequence. 
The point features are then projected onto the 2D feature planes using intrinsic and extrinsic matrices to get semantic information. Further, they are concatenated to form sparse fusion features \(\{\mathcal{F}^s_{\rm fusion}\}^S_{s=1} \in \mathbb{R}^{N \times 2C_s}\). Finally, they are fed into a DETR-like decoder and then finish 9-degree-of-freedom bounding box regression.

However, due to the large number of points in the depth-reconstructed point cloud and computational limitations, only a sparse subset (about 2\%) is sampled, resulting in a negative influence on the point cloud manifold. Due to the varying manifold structures in different local regions of scene-level point clouds, applying a global transformation to the entire point cloud is infeasible. To address this issue, we introduce Proxy Transformation to modify point cloud \textbf{submanifolds} with Proxy Attention, providing richer geometric information for subsequent feature learning.

\subsection{Proxy Attention}
\label{sub:proxy}
For simplicity, we use following formulation to represent Softmax attention:
\begin{equation}
    O = {\rm \sigma}(QK^T)V \triangleq {\rm Attn}(Q,K,V)
\end{equation}
where $ Q, K, V\in\mathbb{R}^{N \times C} $ denote query, key and value matrices and $\rm{\sigma}(\cdot)$ represents Softmax function. And our proxy attention consists of three components as follows.
\begin{equation} \label{eq:attn}
    \begin{split}
        O^{\rm P}&= {\rm ProxyAttn}(Q,K,V,P) \\
        &=\ {\rm \sigma}(QP^T)\ {\rm \sigma}(PK^T)\ V \\ 
        &=\ \underbrace{{\rm Attn}(Q,P,\underbrace{{\rm Attn}(P,K,V)}_{\rm Proxy\ Compression})}_{\rm Proxy\ Broadcast}. \\
    \end{split}
\end{equation}
where \(P \in \mathbb{R}^{n \times C} \) is our proxy tokens.

\noindent \textbf{Proxy Compression.}
Previous works~\cite{han2023agent,pu2024efficient} have shown significant redundancy in query-key similarity computations within self-attention. To address this, we first compress the attention feature space using proxy tokens, whose number is substantially smaller than \(N\) practically. This approach significantly alleviates computational overhead while preserving representational capacity, as shown in~\cref{tab:attn}. Specifically, we use proxy tokens as queries to compute similarity with initial keys, subsequently aggregating original values to obtain a compressed proxy value representation \(V^{\rm P}\) with reduced redundancy.

\noindent \textbf{Proxy Broadcast.}
Subsequently, we treat proxy tokens \(P\) and proxy values \(V^{\rm P}\) as keys and values for a second softmax operation with the original queries, allowing proxy features to be broadcast globally. This design preserves global contextual representation while reducing complexity to a linear scale. In practice, proxy tokens are selected based on the model's representational needs (\eg  downsampling points after pooing, multi-view image features or textual features). In these contexts, \(n\) is much smaller than \(N\). Thus, this approach achieves generalized attention with a linear complexity of $\mathcal{O}(Nnd)$ relative to the number of input features $N$, balancing computational efficiency with high representational capacity as shown in~\cref{tab:attn}.

\noindent \textbf{Proxy Bias.}
Additionally, to compensate for the lack of positional information and the diversity of features, we introduce a novel proxy bias. To minimize parameter overhead, we merge two low-dimensional subspaces into a higher-dimensional space through a union operation of linear space. Specifically, we define two sets of low-dimensional learnable parameters, representing two subspaces $V_c$ and $V_r$ within feature space $V$ that satisfy $V = V_c \cup V_r$. After broadcasting these two parameter sets, they match the original feature dimensionality, enriching the feature space with positional information, and guiding Proxy Attention to focus on diverse regions.

\subsection{Deformable Point Clustering}
\label{sub:deform}
To transform submanifolds, it is essential first to obtain suitable 
 clusters. Motivated by \cite{deformabledetr,dai2017deformable,xia2022vision}, we design a deformable clustering approach to identify optimal cluster center locations, capturing the desired target submanifolds.

\noindent \textbf{Grid Prior.}
As mentioned in \cref{sub:overall}, the randomly sampled sparse point cloud has already lost a significant amount of geometric information. To compensate for this loss, we first generate a 3D uniform grid and use its grid centers as reference points for the initial clustering:
\begin{equation}\label{eq:grid}
    \begin{aligned}
        \mathcal{Q}\  &\triangleq \{q_{(i,j,k)}\}_{i,j,k=1}^{x_s,y_s,z_s} \sim U[\mathcal{C}_{min},\mathcal{C}_{max}], \\
        \mathcal{N}_{(i,j,k)} &\triangleq \{\ p_l^{i,j,k}\}_{l=1}^{m} = \gamma\ (\ \mathcal{Q}_{(i,j,k)},\ \{p_l\}_{l=1}^N\ ),
\end{aligned}
\end{equation}
where reference points $\mathcal{Q}$ are also the initial cluster centers. $U$ is a uniform distribution. $\mathcal{C}$ is the 3D coordinate of points. $p_l^{i,j,k}$ is a specific point in corresponding cluster $\mathcal{N}_{(i,j,k)}$ and $\gamma$ is clustering function, \eg ball query or kNN. $N$ and $m$ are respective numbers of point cloud and local cluster.

To simplify, we regard $\mathcal{N}_{(i,j,k)}$ as $\mathcal{N}_t$, where $t \in [0, n-1]$, where $n=x_s \times y_s \times z_s$ and $x_s,y_s,z_s$ are predefined as hyperparameter. Experimentally, we find that this approach does not affect performance, but provides greater stability in the training process compared to stochastic sampling.

\noindent \textbf{Deformable Offsets.}
To increase the diversity of submanifold selection, an offset network outputs the offset values for reference points respectively, allowing them to shift toward more relevant surrounding regions based on the current submanifold structure. Considering that each reference point covers a local $s$ ball region ($s$ is the largest value for offset), the offset network is able to percept local features to learn reasonable offsets. In addition, we also provide global information to further enhance offset generation. 
\begin{equation}\label{eq:offset}
    \begin{aligned}
        \hat{q}_{t} &= q_{t} + \Gamma_{\rm{offset}}(\mathcal{N}_t), \\
        \hat{\mathcal{N}}_t &= \gamma\ (\ \hat{q_{t}},\ \{\ p_l\}_{l=1}^N\ ),
    \end{aligned}
\end{equation}
where $\Gamma_{\rm{offset}}(\cdot)$ is our offset network. $\hat{q}_t \in \hat{\mathcal{Q}}$ is a new cluster center and $\hat{\mathcal{N}}_t$ is a new neighborhood.

Specifically, for $\mathcal{N}_t$, we first concatenate the relative coordinates of sub-points with respect to the center and their global coordinates. It's then passed through $1 \times 1$ Conv2D and ReLU activation to extract point-wise features. We apply average pooling to obtain high-dimensional local-global features for each cluster, which are then fed into another $1 \times 1$ convolution to generate the final offsets, as shown in~\cref{fig:framework}. Notably, we set the bias of the $1 \times 1$ convolution to zero to avoid introducing additional perturbations. Finally, we discard some clusters according to a predefined drop ratio $\beta$ with a certain sampling method(\eg random sampling or FPS), allowing us to obtain a more diverse neighborhood based on the new cluster centers.

\subsection{Proxy Transformation}
\label{sub:trans}
For each submanifold, we leverage the Proxy Block to learn a unique manifold transformation. Any arbitrary transformation in 3D space can be decomposed into a linear transformation and a translation. Then, these transformations are applied to each deformable cluster. Details about the transformation generation module are described below.

\noindent \textbf{Proxy Block.} Following the standard block design in Transformer architectures~\cite{vaswani2017attention}, we developed a Proxy Block based on proxy attention, aimed at cross-modal learning for subsequent manifold transformations. We denote $P_0$ and $F_0$ as a set of proxy tokens and cluster features. And $F_0$ is extracted by a simplified PointNet~\cite{qi2017pointnet}. Then a transformer block comprises of proxy attention module and feedforward network:
\begin{equation}\label{b1}
    F = F_0 + B^{\rm P},
\end{equation}
\begin{equation}\label{b2}
    Q = FW_Q, K = FW_K, V = FW_V, P = P_0W_P,
\end{equation}
\begin{equation}\label{b3}
    O^{\rm P} = F + {\rm ProxyAttn}(Q,K,V,P),
\end{equation}
\begin{equation}\label{b4}
    O = O^{\rm P} + {\rm FFN}(O^{\rm P}),
\end{equation}
where $B^{\rm P}$ is our proxy bias introduced in \cref{sub:proxy}. $W_Q,W_K,W_V$ are are projections for query, key and value. ${\rm ProxyAttn}(\cdot)$ is a proxy attention module in \cref{eq:attn}. ${\rm FFN}(\cdot)$ is a position-wise feedforward network.

In the following sections, for simplicity, we use
\begin{equation}
    O = {\rm ProxyBlock}(F, P),
\end{equation}
to represent the basic proxy block (\cref{b1,b2,b3,b4}).

\noindent \textbf{Text Guided Translation.}
In ego-centric 3D visual grounding, the input text contains global relative positional information between the target object and other reference objects, while the remaining parts naturally correspond to background point clouds. Therefore, text features can effectively guide the model in learning translation vectors that establish spatial relationships between various submanifolds, thereby enhancing the global relational information embedded within the point cloud structure. Specifically, we leverage text features as proxies within the Proxy Block, allowing it to generate text-guided point cloud features that ultimately yield accurate and context-aware final translation vectors for more precise grounding results.
\begin{equation}
    F^{l+1} = {\rm ProxyBlock}(F^l, P_{T}),l=0,..,L-1, 
\end{equation}
\begin{equation}
    T = F^{L}U_{\rm text},
\end{equation}
where $P_T$ are text features $\mathcal{F}_{\rm text} \in \mathbb{R}^{S \times C}$. $L$ is number of layers and $U_{\rm text}$ is a fully connected layer to generate final translation vectors $T \in \mathbb{R}^{n \times 3}$.

\noindent \textbf{Image Guided Transformation.}
The input multi-view image contains rich fine-grained semantic information, compensating for the contextual loss caused by point cloud downsampling. Multi-view images provide pose information of the target object from various perspectives, guiding the model to learn internal transformations of each sub-points relative to the cluster center. Thus, we use pooled image features as proxies in the Proxy Block to generate reshaped image-guided point cloud features, which in turn yield the linear transformations for each submanifold:
\begin{equation}
    F^{l+1} = {\rm ProxyBlock}(F^l, P_{I}),l=0,..,L-1, 
\end{equation}
\begin{equation}
    M = F^{L}U_{\rm image},
\end{equation}
where $P_I \in \mathbb{R}^{V \times C}$ are pooled image features from an additional attention pooling layer. And $L$ is the number of layers and $U_{\rm image}$ is a fully connected layer to generate final transformation matrices $M \in \mathbb{R}^{n \times 3 \times 3}$.

\noindent \textbf{Submanifold Reshape.}
Once the transformation matrixs, containing fine-grained semantic information from the image, and the translation vectors, encompassing precise global relative information from the text, are obtained, we can apply a Proxy Transformation to each submanifold and finally get a diverse enhanced point cloud manifold. To represent this process, we use  concise set operations:
\begin{equation}\label{eq:set trans}
    \hat{\mathcal{P}} = \mathcal{M} \odot \mathcal{P}\ \  \oplus \ \ \mathcal{T}, 
\end{equation}
where $\mathcal{M}$ and $\mathcal{T}$ are sets of transformation matrixs and translation vectors for all clusters. $\mathcal{P}$ is a union set of submanifolds of different neighborhoods $\{{\mathcal{N}_i}\}_{i=1}^n$, namely a subset of original point cloud. $\odot$ and $\oplus$ represent element-wise multiplication and addition for sets.

Equivalently, for a specific submanifold:
\begin{equation}\label{eq:element trans}
    \hat{P}_i = P_iM_i^T + T_i,
\end{equation}
where $P_i \in \mathcal{P} \cap \mathbb{R}^{m \times 3}$ represent sub-points in $\mathcal{N}_i$. $M_i \in \mathcal{M} \cap \mathbb{R}^{3 \times3}$ and $T_i \in \mathcal{T} \cap \mathbb{R}^{3}$ are the specific transformation matrix and translation vector for this submanifold.

Finally, we replace the corresponding parts of the original point cloud with transformed submanifolds to obtain final enhanced point cloud, consisting of rich diversity in manifold structure, which is then fed into 3D backbone during training, enabling efficient multimodal-guided point cloud manifold enhancements.

%% file: sec/5_experiment.tex
\section{Experiments}
\label{sec:experiments}
\begin{table*}
    \centering
    \begin{tabular}{c|c@{\hskip 4pt}c|c@{\hskip 2pt}c|c|c@{\hskip 4pt}c|c@{\hskip 2pt}c|c}
    \toprule
    \multirow{2}{*}{Method}  & Easy & Hard & Indep & Dep & Overall & Easy & Hard & Indep & Dep & Overall\\
    & AP$_{25}$ & AP$_{25}$ & AP$_{25}$ & AP$_{25}$ & AP$_{25}$& AP$_{50}$ & AP$_{50}$ & AP$_{50}$ & AP$_{50}$ & AP$_{50}$ \\
    \midrule
    ScanRefer$^\dagger$~\cite{chen2020scanrefer}  & 13.78& 9.12& 13.44& 10.77&12.85 &- & -& -&- & -\\
    BUTD-DETR$^\dagger$~\cite{jain2022bottom}  & 23.12& 18.23& 22.47& 20.98&22.14 & -& - & - & - & -\\
    L3Det$^\dagger$~\cite{zhu2023object2sceneputtingobjectscontext}  & 24.01& 18.34& 23.59& 21.22&23.07 & - & - & - & - &- \\
    EmbodiedScan$^\dagger$  & 39.82 & 31.02 & 40.30 & 38.48 & 39.10 & 18.79  & 14.93  &  18.03 &  18.71 & 18.48\\

    EmbodiedScan~\cite{wang2023embodiedscan}  & 33.87& 30.49& 33.55& 33.61&33.59 & 14.58& 12.41& 13.92& 14.65&14.40\\

    DenseG~\cite{zhengdenseg} &40.17 &34.38 &38.79 &40.18 &39.70 &18.52 &\textbf{15.88} &17.47 &18.75 &18.31 \\

    \textbf{ProxyTransformation} &
    \textbf{41.66}   &
    \textbf{34.38}  &
    \textbf{41.57} &
    \textbf{40.81} & 
    \textbf{41.08}
    &
    \textbf{19.43}  &
    14.09  &
    \textbf{18.65} &
    \textbf{19.18} & 
    \textbf{19.00}\\

    \bottomrule
    \end{tabular}
    \caption{Main Results on the official validation set. The table displays accuracy performance considering instances where IoU exceeds 25\% and 50\% under different circumstance as mentioned in~\cref{sec:experiments}. 'Indep' and 'Dep' mean the targets are view-independent and view-dependent. And $^\dagger$ denotes that models are trained on full train dataset. Although our model is trained on mini train dataset (approximately 20\% of the full data), but it still surpasses previous methods trained on the full train dataset.}
    \label{table:val}
\end{table*}

\noindent\textbf{Dataset and Benchmark.}
The EmbodiedScan dataset~\citep{wang2023embodiedscan} used in our experiments is a large-scale, multi-modal, ego-centric resource tailored for comprehensive 3D scene understanding. It comprises \textbf{5,185} scene scans from widely-used datasets like ScanNet~\citep{dai2017scannet}, 3RScan~\citep{wald2019rio}, and Matterport3D~\citep{chang2017matterport3d}. This diverse dataset offers a rich foundation for 3D visual grounding, covering a broader range of scenes than previous datasets. The training set includes \textbf{801,711} language prompts, while the validation set contains \textbf{168,322} prompts, making EmbodiedScan notably larger and more challenging, thus providing a rigorous benchmark for ego-centric 3D visual grounding tasks.

\noindent\textbf{Experimental Settings.}
Due to limited resources, we train on the official mini dataset. Despite this, our model outperforms the official baseline trained on the full training set, as shown in~\cref{table:val}. For analysis experiments in~\cref{sub:analysis}, we also use the mini training and validation sets available through the official EmbodiedScan release~\citep{wang2023embodiedscan}.

Our reported accuracy uses the official IoU metric, focusing on cases where IoU exceeds 25\% and 50\%. We also assess model performance on both "Easy" and "Hard" scenes, where a "Hard" scene contains three or more target objects of the same category. The "View-Dependent" and "View-Independent" metrics further test the model’s spatial reasoning by evaluating performance with and without perspective-specific descriptions.

\noindent\textbf{Implementation Details.}
This work builds upon the strong baseline from EmbodiedScan for ego-centric 3D visual grounding. We applied several techniques to enhance the original baseline. First, we replace the RoBERTa language encoder~\citep{liu2019robertarobustlyoptimizedbert} with the CLIP encoder~\citep{radford2021learning} to achieve superior alignment between language and vision. Additionally, we incorporate Class-Balanced Grouping and Sampling (CBGS)~\cite{zhu2019class} during pretraining to address data imbalance and enhance detection accuracy across rare and common object categories. We follow EmbodiedScan’s approach by using pretrained visual encoders. Finally, we also adopt text augmented by LLM to provide richer semantic context when training, proposed by~\cite{zhengdenseg}. Moreover, our Proxy Transformation is trained using the AdamW optimizer with a learning rate of 5e-4, weight decay of 5e-4, and a batch size of 48. Training spans 12 epochs, with the learning rate reduced by 0.1 at epochs 8 and 11.

\subsection{Main Results}\label{sub:main results}
We evaluate the 3D visual grounding performance of our proposed method, Proxy Transformation, with results presented in~\cref{table:val}, comparing it against established SOTA methods from the dataset benchmark. In addition, the upper methods with $^\dagger$ are trained on the full dataset, while the lower methods are trained on a mini dataset, approximately 20\% of the full dataset.

As shown in~\cref{table:val}, Proxy Transformation achieves a notable 7.49\% and 4.60\% improvement over the previous strongest baseline, EmbodiedScan, on the Overall@0.25 and Overall@0.50. Although only trained on the mini dataset, our method even surpasses the baseline trained on the full dataset. Through deformable point clustering, our model focuses on the most crucial target regions in the scene, reducing the extra computation overhead caused by redundant point clouds and improving efficiency. Additionally, a grid prior preserves essential original spatial information, mitigating early training instability. Recognizing that text information provides global positional relationships among different submanifolds and image information offers local semantic details within each submanifold, we designed generalized proxy attention to guide local transformations of the selected point cloud submanifolds. These transformations optimize local point cloud structures for each submanifold, ultimately providing higher-quality data for subsequent feature extraction and fusion.

In summary, \textbf{Proxy Transformation} consistently outperforms the previous state-of-the-art across multiple evaluation metrics, including easy and hard samples as well as view-independent and view-dependent tasks. These consistent improvements across diverse metrics highlight not only the robustness and generalizability of our approach for 3D visual grounding tasks but also its adaptability to complex scenes and dynamic environments, making it an effective solution for advancing multimodal grounding capabilities in real-world embodied AI applications.

\subsection{Analysis Experiments}
\label{sub:analysis}

\begin{table}
    \centering
    \begin{tabular}{c c c|c c |c}
    \toprule
    Grid Prior & Offsets & PT & Easy& Hard& Overall\\
    \midrule
    &&&37.05 & 30.60 & 36.53 \\
     & &\checkmark & 40.39 & 32.60 & 39.76 \\
    &\checkmark&\checkmark & 40.59 & 32.18 & 39.91 \\
    \checkmark &\checkmark & \checkmark & \textbf{41.66} & \textbf{34.38} & \textbf{41.08}\\
    \bottomrule
    \end{tabular}
    \caption{Ablation of Proposed Modules. 'PT' represents 'Proxy Transformation'. The reported values are mAP for predictions greater than 25\% IoU.}
    \label{tab:module ablation}
\end{table}

\begin{table}
    \centering
    \begin{tabular}{c |c @{\hskip 4pt}c |c @{\hskip 4pt}c |c}
    \toprule
    Attn & FLOPs & \#Params & Easy &  Hard& Overall\\
    \midrule
     Self & 8.36G & 2.52M & 40.71 & 33.65 & 40.14 \\
     Cross & 4.53G & 2.42M & 36.87 & 28.81 & 36.22 \\
     Proxy   & \textbf{4.97G}&2.71M & \textbf{41.66} & \textbf{34.38} & \textbf{41.08} \\
    \bottomrule
    \end{tabular}    
    \caption{Ablation on Proxy Attention. Here, we compare the vanilla self-attention block and cross-attention block with our Proxy Block. The FLOPs and parameters are computed over 3 transformer blocks with cluster features as input. We use a grid size of $12$ and a drop radio of $0.6$. The reported values are mAP for predictions greater than 25\% IoU.}
    \label{tab:attn}
\end{table}

\begin{table}
    \centering
    \begin{tabular}{c |c c |c}
    \toprule
    Grid Size & Easy &  Hard& Overall\\
    \midrule
     w/o & 40.59 & 32.18 & 39.91 \\
     10 & 40.94 & 32.81 & 40.29 \\
     \textbf{12} & \textbf{41.66} & \textbf{34.38} & \textbf{41.08} \\
     14 & 41.11 & 34.28 & 40.56 \\
    \bottomrule
    \end{tabular}
    \caption{Ablation on the size of Grid Prior. The reported values are mAP for predictions greater than 25\% IoU.}
    \label{tab:grid}
\end{table}

\begin{figure*}
    \centering
    \includegraphics[width=\linewidth]{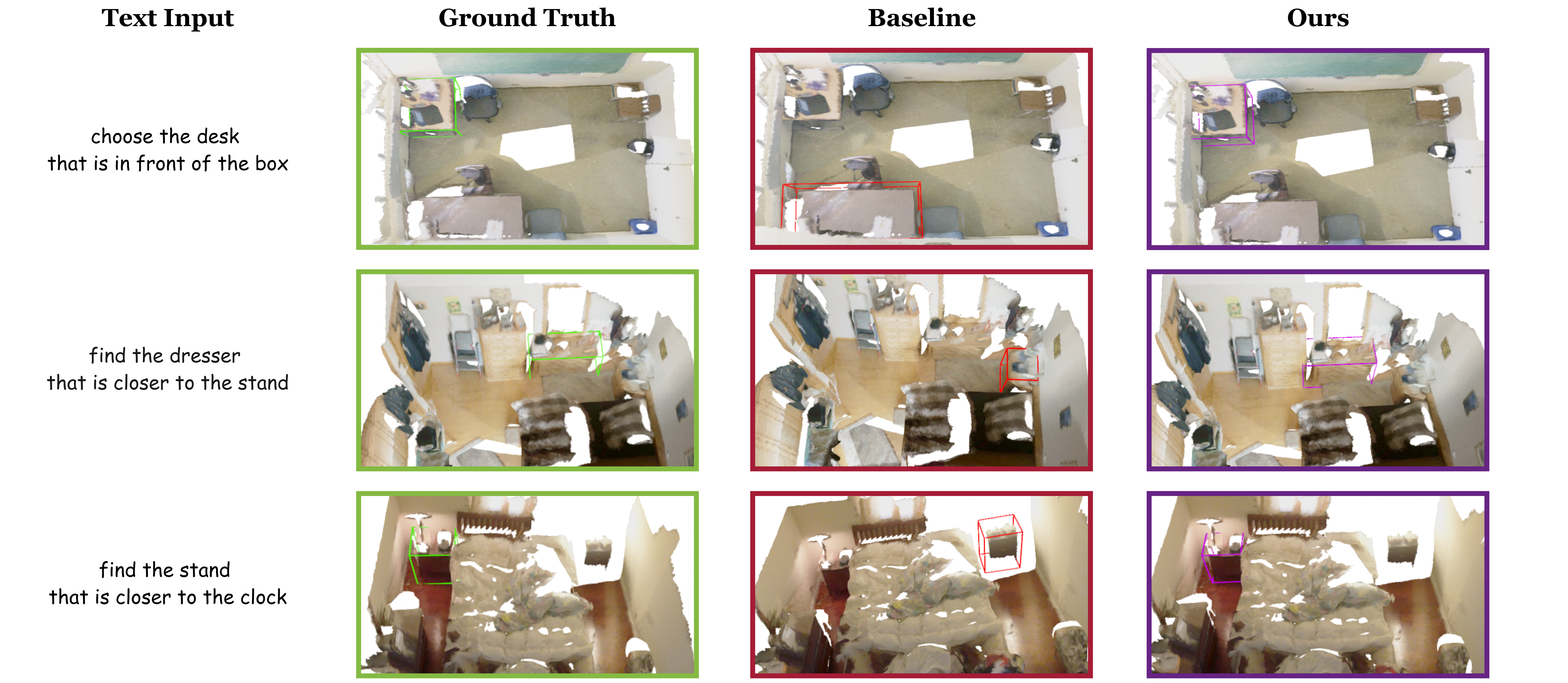}
    \caption{Visualization of ground truth and predictions. Ground truth boxes are shown in \textcolor{green}{green}, baseline in \textcolor{red}{red}, and ProxyTransformation's predictions in \textcolor{violet}{violet}.}
    \label{fig:predict_anchors}
    \vspace{-0.2cm}
\end{figure*}


\noindent \textbf{Ablation on Proposed Modules.}
We conducted ablation studies on each module to verify their effectiveness as shown in~\cref{tab:module ablation}. To ensure fair comparisons, the baseline is enhanced with several modules described in the implementation details. First, we introduce Proxy Transformation without deformable point clustering, resulting in a substantial improvement of 3.23\%, which demonstrates the effectiveness and necessity of our multimodal point cloud enhancement. Then, we add deformable offsets, yielding an additional improvement on easy samples, indicating that offsets effectively guide cluster centers toward target regions. Finally, we incorporate grid prior, which improves training stability and achieves an increase of 2.20\% on hard samples. This result shows that the grid prior mitigates the randomness of offsets in the early stages of training, preserving essential scene information for the model.

\noindent \textbf{Ablation on Proxy Attention.}
We validate the effectiveness and efficiency of proxy attention by comparing the vanilla attention block with the proxy block. The input cluster features are aligned across both blocks. As discussed in~\cref{sub:proxy}, the number of proxy tokens is significantly smaller than the input sequence length. This reduction in proxy tokens compresses redundancy within the attention module, enhancing computational efficiency, which aligns with our experiment results. As shown in~\cref{tab:attn}, our approach achieves higher performance than vanilla attention, while reducing the computation overhead by \textbf{40.6\%}. And we have added an additional baseline that applies cross-attention as shown in \cref{tab:attn}, which demonstrate that our proposed Proxy Attention achieves a better trade-off between accuracy performance and computational efficiency.

\begin{table}
    \centering
    \begin{tabular}{c |c c |c}
    \toprule
    Drop Radio & Easy &  Hard& Overall\\
    \midrule
     0.5 & 40.55 & 34.17 & 40.03 \\
     \textbf{0.6} & \textbf{41.66} & \textbf{34.38} & \textbf{41.08} \\
     0.7 & 40.67 & 32.70 & 40.02 \\
     0.8 & 39.65 & 31.02 & 38.95 \\
     0.9 & 36.40 & 27.02 & 35.64 \\
    \bottomrule
    \end{tabular}
    \caption{Ablation on Cluster Drop Radio $\beta$. A smaller \( \beta \) indicates that a greater number of submanifolds are used for subsequent transformations. The reported values are mAP for predictions greater than 25\% IoU.}
    \label{tab:drop radio}
\end{table}

\noindent \textbf{Ablation on Main Hyperparameters.} 
Then, we ablate several hyperparameters in deformable point clustering and proxy transformation. As shown in~\cref{tab:grid}, the size of the grid prior also influences performance by reducing instability during the early training stages, which can result from the randomness of the initial transformations.
As shown in~\cref{tab:drop radio}, an excessively high drop ratio reduces model performance, highlighting the effectiveness of our submanifold transformations. A lower drop ratio retains more submanifolds for subsequent Proxy Transformation, enhancing results. However, experiments show that an overly low drop ratio also degrades performance, as point cloud enhancement should complement rather than overly alter the original structure. Moreover, further ablation studies are given in the supplementary material.

\subsection{Qualitative Results and Discussion}
\label{sub:qualitative}
With our deformable point clustering and Proxy Transformation, the point cloud structure in target regions is optimized, providing higher-quality data for subsequent feature learning and fusion. As shown in~\cref{fig:predict_anchors}, reference objects in these regions are small and difficult to distinguish, but with the enhanced manifold structure, our model effectively captures the relationships between target and reference objects, achieving improved grounding performance. 

%% file: sec/6_conclusion.tex
\section{Conclusion}
In this work, we propose Proxy Transformation, a real-time multimodal point cloud enhancement model, to address issues of geometric loss and unexpected noise in 3D points within ego-centric 3D visual grounding. This approach yields a high-quality point cloud structure for feature learning. Extensive experiments show that our model not only outperforms all existing methods but also improves the computational efficiency of attention modules, setting a new SOTA on the EmbodiedScan benchmark.
These results demonstrate the effectiveness and robustness of using multimodal information for real-time point cloud enhancement in ego-centric 3D visual grounding, offering fresh insights for advancing embodied AI and enabling human-computer interaction in real-world environments.


%% file: sec/X_suppl.tex
\clearpage

\section*{{\large Appendix}}


\section*{A. Additional Experiments}

\noindent \textbf{Ablation on the number of Proxy Layers.}
As shown in~\cref{tab:layers}, we conduct several experiments to determine the optimal layers of our proxy attention block for efficient learning. Based on these findings, we selected the optimal hyperparameters for our experiments. But we can see that even with just a single layer our model can achieve a SOTA performance. This demonstrates that the gain stems from leveraging multi-modal information rather than merely increasing model parameters. Thus, we can use a single-layer as final setup in practice, achieving SOTA results with minimal computational cost.

\noindent \textbf{Ablation on $s$.}
As discussed when introducing deformable offsets, \(s\) represents the maximum value of the offsets. Before generating the grid prior, we scale down the cuboid defined by the maximum and minimum coordinates of the point cloud based on \(s\), ensuring that reference points, when adjusted by deformable offsets, do not move outside the boundaries of the point cloud. As shown in~\cref{tab:max value}, we selected an optimal maximum value for \(s\). An excessively large \(s\) results in the reduced preset grid losing essential prior information, while an overly small \(s\) prevents reference points from shifting toward more critical target regions, thereby reducing the flexibility of the model.

\section*{B. Ego-Centric 3D Visual Grounding}

In real-world applications, intelligent agents interact with their surroundings without prior knowledge of the entire scene. Instead of relying on pre-reconstructed 3D point clouds or other scene-level priors commonly used in previous studies~\citep{wu2024pointv3, huang2023segment3d}, they primarily depend on ego-centric observations, such as multi-view RGB-D images.

Following the definition in \cite{wang2023embodiedscan}, we formalize the ego-centric 3D visual grounding task as follows: Given a natural language query \(L \in \mathbb{R}^T\), along with \(V\) RGB-D image views \(\{(I_v, D_v)\}_{v=1}^V\), where \(I_v \in \mathbb{R}^{H \times W \times 3}\) denotes the RGB image and \(D_v \in \mathbb{R}^{H \times W}\) represents the depth map for the \(v\)-th view, and their corresponding sensor intrinsics \(\{(K^I_v, K^D_v)\}_{v=1}^V\) and extrinsics \(\{(T^I_v, T^D_v)\}_{v=1}^V\), the goal is to predict a 9-degree-of-freedom (9DoF) bounding box \(B = (x, y, z, l, w, h, \theta, \phi, \psi)\). 

In this context, \((x, y, z)\) specify the 3D center coordinates of the target object, \((l, w, h)\) define its dimensions, and \((\theta, \phi, \psi)\) represent its orientation angles. The task is to determine \(B\) such that it accurately localizes the object described by \(L\) within the scene captured by \(\{(I_v, D_v)\}_{v=1}^V\).

\section*{C. Details about Proxy Bias}
As mentioned in the methodology, to compensate for the lack of positional information and the diversity of features, we propose a novel \textbf{Proxy Bias}:
\begin{equation}
    F = F_0 + B^{\rm P},
\end{equation}
where $F \in \mathbb{R}^{N \times C}$ is the input of Proxy Block and $F_0 \in \mathbb{R}^{N \times C}$ is our deformable cluster features. $B^{\rm P} \in \mathbb{R}^{N \times C}$ is our novel proxy bias.

Initially, we set three learnable parameters $B_d \in \mathbb{R}^{N \times D \times D}$, $B_c \in \mathbb{R}^{N \times 1 \times S}$ and $B_r \in \mathbb{R}^{N \times S \times 1}$. Here, $C = S^2 = D^4$. Therefore, our parameters are way less than directly setting $B^{\rm P}$ as a learnable parameter, thus improving our parameter efficiency.

We first interpolate \( B_d \) into \( B_1 \in \mathbb{R}^{N \times S \times S} \), mapping the low-dimensional subspace into a higher-dimensional feature space to enhance feature diversity. Subsequently, we add \( B_c \) and \( B_r \) to obtain the final \( B_2 \in \mathbb{R}^{N \times S \times S} \), representing the linear union of two low-dimensional subspaces to form the final high-dimensional space, expressed as \( V = V_1 \cup V_2 \). Finally, we get $B^{\rm P} = (B_1 + B_2).{\rm reshape}(N,C)$, which can enrich the feature space with positional information and guide ProxyAttention to focus on diverse regions.

\begin{table}
    \centering
    \begin{tabular}{c |c c |c}
    \toprule
    Max & Easy &  Hard& Overall\\
    \midrule
     3 & 38.14 & 29.65 & 37.46 \\
     \textbf{4} & \textbf{41.66} & \textbf{34.38} & \textbf{41.08} \\
     5 & 36.67 & 26.71 & 35.87 \\
    \bottomrule
    \end{tabular}
    \caption{Ablation on the Max Value of Offsets. The reported values are mAP for predictions greater than 25\% IoU.}
    \label{tab:max value}
\end{table}

\begin{table}
    \centering
    \begin{tabular}{c |c c |c}
    \toprule
    Layers & Easy &  Hard& Overall\\
    \midrule
     1 & 41.60 & 34.07 & 40.99 \\
     2 & 41.57 & 31.97 & 40.79 \\
     3 & 41.66 & \textbf{34.38} & \textbf{41.08} \\
     4 & \textbf{41.72} & 31.55 & 40.90 \\
    \bottomrule
    \end{tabular}
    \caption{Ablation on the number of Proxy Layers. The reported values are mAP for predictions greater than 25\% IoU.}
    \label{tab:layers}
    \vspace{-0.2cm}
\end{table}